\documentclass[letterpaper]{article}
\usepackage{natbib,alifeconf,todonotes,enumitem,array,amsmath,amssymb,layouts,mathtools, booktabs, url, amssymb, float, placeins}

\title{Systematic Derivation of Behaviour Characterisations in Evolutionary Robotics}
\author{Jorge Gomes$^{1,2}$ \and Pedro Mariano$^2$ \and Anders Lyhne Christensen$^{1,3}$ \\
\mbox{}\\
$^1$Instituto de Telecomunica\c{c}\~{o}es, Lisbon, Portugal \\
$^2$LabMAg -- Faculdade de Ci\^{e}ncias da Universidade de Lisboa, Portugal \\
$^3$Instituto Universit\'{a}rio de Lisboa (ISCTE-IUL), Lisbon, Portugal \\
\texttt{jgomes@di.fc.ul.pt, plmariano@fc.ul.pt, anders.christensen@iscte.pt}}

\newcolumntype{L}[1]{>{\raggedright\let\newline\\\arraybackslash\hspace{0pt}}m{#1}}
\newcolumntype{C}[1]{>{\centering\let\newline\\\arraybackslash\hspace{0pt}}m{#1}}
\newcolumntype{R}[1]{>{\raggedleft\let\newline\\\arraybackslash\hspace{0pt}}m{#1}}

\begin{document}
\maketitle

\begin{abstract}
Evolutionary techniques driven by behavioural diversity, such as novelty search, have shown significant potential in evolutionary robotics. These techniques rely on priorly specified behaviour characterisations to estimate the similarity between individuals. Characterisations are typically defined in an \emph{ad hoc} manner based on the experimenter's intuition and knowledge about the task. Alternatively, generic characterisations based on the sensor-effector values of the agents are used. In this paper, we propose a novel approach that allows for systematic derivation of behaviour characterisations for evolutionary robotics, based on a formal description of the agents and their environment. Systematically derived behaviour characterisations (SDBCs) go beyond generic characterisations in that they can contain task-specific features related to the internal state of the agents, environmental features, and relations between them. We evaluate SDBCs with novelty search in three simulated collective robotics tasks. Our results show that SDBCs yield a performance comparable to the task-specific characterisations, in terms of both solution quality and behaviour space exploration.
\end{abstract}



\section{Introduction}
Evolutionary robotics (ER) is focused on the application of evolutionary algorithms to the synthesis of robot controllers, and sometimes robot morphologies \citep{nolfi00}. In ER, the evolutionary process is typically driven towards solutions for a task according to a manually crafted fitness function \citep{nelson09}. Evolutionary algorithms driven by a fitness function are, however, vulnerable to deception and convergence to local optima \citep{whitley91}. Common approaches to overcome these issues are based on diversity preservation. Traditionally, diversity is preserved at the genomic level \citep{goldberg87}. However, in ER it has been observed that many different genotypes can represent similar behaviours, and that small changes in the genotype can lead to substantial changes in behaviour \citep{mouret12}. Therefore, recent works have proposed the use of behaviour similarity measures (BSM) in the evolutionary process to promote an effective diversity in the population.

A number of different strategies using BSM have been proposed. The most popular approach, novelty search, directly rewards individuals that display novel behaviours (as measured by the BSM), instead of using a fitness function \citep{lehman11ns,mouret12}. BSM have also been used to speciate the population according to the behaviour of the individuals \citep{trujillo11,moriguchi10}. \citet{ollion11} defined a measure for the degree of behaviour space exploration, based on BSM, and used the measure to predict the relative performance of different evolutionary setups.

Devising effective behaviour measures is not straightforward \citep{savage04}, since the measure must be able to capture the details of the behaviour that are relevant in the given task, while at the same time should contain no or as few irrelevant behaviour features as possible. Behaviour measures are typically defined by the experimenter based on task-specific knowledge (for examples, see \citealp{mouret12,lehman11ns,mouret09,gomes13si}). Relying on the experimenter's intuition about the task may, however, introduce biases in the evolutionary process. These biases can compromise the evolutionary search, as the experimenter might not know beforehand what type of behavioural traits should be explored in order to solve the task.


Several researchers have proposed the use of \emph{generic} BSM \citep{doncieux10,gomez09,gomes13gecco} for ER. Generic measures do not rely on the specific details of a task, and can be used throughout the domain. Their use is, however, associated with two key issues: (i)~generic measures are only weakly related with the specific task, which can create an unnecessarily large behaviour space \citep{cuccu11}, and (ii)~they are mostly unintelligible, undermining analysis and comprehension of the behaviour space exploration. 

We propose a new class of behaviour measures, \emph{systematically derived behaviour characterisations} (SDBCs), that combine the advantages of task-specific and generic measures. The proposed measures are directly derived from a formal description of the task state. This way, we reduce the dependency on the experimenter's knowledge about the task while, at the same time, obtain characterisations that are directly related to the task. In this paper, we apply SDBCs to robotics tasks, but the approach could potentially be generalised to other types of agent-based systems, as the formal task description resembles the \textsc{SMART} agent framework \citep{inverno04}. We augment our method with the inclusion of a weighting scheme for behaviour features, based on the estimated relevance of the features.


The proposed measures are evaluated with novelty search in three simulated collective robotics tasks: resource sharing, gate escape, and predators-prey pursuit. We compare the proposed measures with behaviour characterisations defined based on task-specific knowledge.

\section{Related Work}

%


\subsection{Task-specific distance measures}
\label{sec:ts}

Most previous works on behavioural diversity rely on behaviour characterisations designed specifically for the given task. These characterisations are composed of behavioural traits that the experimenter considers relevant for describing agent behaviour in the context of the given task. The behavioural distance then corresponds to the Euclidean distance between the characterisation vectors. Table~\ref{tab:related} lists several examples of characterisations that have been used in previous ER studies.  Analysing these characterisations, the following commonalities can be identified:
\begin{itemize}[noitemsep]
\item There is a strong focus on the spatial relationships between entities in the task environment (1, 2, 5, 6, 8, 9).
\item Characterisations often comprise only a small number of different behavioural traits (1, 3--5, 7--9). 
\item Many characterisations focus on the final state of the environment (1, 3--5, 7--9), values averaged over an entire trial (7--9), or a single quantity sampled over time (2, 6).
\end{itemize}

\begin{table*}[tb]
\caption{Behaviour characterisations used in previous works in evolutionary robotics. The right-most column lists the lengths of the characterisation vectors, with \emph{S} denoting that the length is proportional to the simulation length.}
\begin{center}
\includegraphics[width=\linewidth]{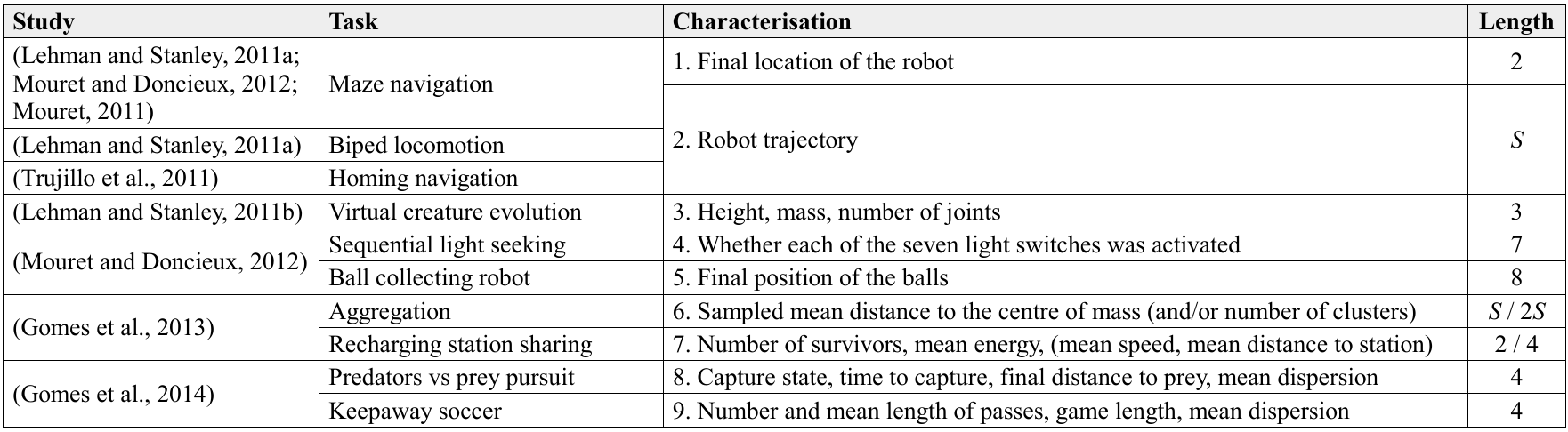}
\vspace{-20pt}
\end{center}
\label{tab:related}
\end{table*}

\subsection{Generic distance measures}

\citet{gomez09} proposed the use of generic measures for assessing behaviour similarity. In the proposed approach, action records for the agents are compared, using either the Hamming distance, relative entropy, or normalised compression distance (NCD). \citet{doncieux10} extended generic measures to evolutionary robotics, proposing the following BSMs:
\begin{description}[noitemsep,font=\normalfont\itshape]
\item[Hamming distance:] Distance between the sequence of all the binary sensor and effector values of the agent sampled through time.
\item[DFT:] A discrete Fourier transform is applied to the sensor-effector sequence, and the first coefficients are used to compute the distances between individuals.
\item[State count:] Each possible sensor-effector state corresponds to one entry in a vector. Each entry contains the number of times the corresponding state was visited.
\end{description}
\citet{gomes13gecco} extended the \emph{Hamming distance} and \emph{state count} measures, making them applicable to multiagent systems and non-binary sensors and effectors.

Common to all generic measures is that they rely exclusively on the sensor-effector states of the agents. While this design makes the measures generic and widely applicable, as any robotics experiment involves robots with sensors and effectors, it also represents their main weakness. Since there is no direct relationship between a given task and the generic behaviour characterisation, the behaviour space is typically very large, which can compromise the performance of diversity-based techniques \citep{cuccu11,mouret11}. Additionally, the characterisations are mostly unintelligible, since they are not intrinsically related to observable characteristics of the agent's behaviour.

\citet{mouret12} compared task-specific and generic characterisations in a comprehensive empirical study with a number of single-robot tasks. \citet{doncieux13} showed how different similarity measures (generic or task-specific) can be combined, by either switching between them throughout evolution or by calculating the behaviour distance based on all similarity measures.

\subsection{Novelty search}

To evaluate the proposed SDBCs, we use novelty search to evolve controllers for three distinct multirobot tasks. In novelty search \citep{lehman11ns}, individuals are scored according to a \emph{novelty metric}, instead of how well they perform a given task.

The novelty metric quantifies how different an individual is from other, previously evaluated individuals with respect to behaviour. The metric relies on the mean behaviour distance (as given by the BSM) of that individual to the $k$~nearest neighbours. Potential neighbours include the other individuals of the current population and a sample of individuals from previous generations stored in an archive. Candidates from sparse regions of the behaviour space therefore tend to receive higher novelty scores, generating a constant evolutionary pressure towards behavioural innovation.

As novelty search is guided by behavioural innovation alone, its performance can be greatly affected by the size and shape of the behaviour space. In particular, behaviour spaces that are vast or contain dimensions not related with the task can cause novelty search to perform poorly \citep{cuccu11,mouret11,gomes12}. To address this issue, we augment novelty search and include a fitness objective, as suggested in \citep{mouret11,mouret12}.

\section{Systematically Derived\\Behaviour Characterisations -- SDBC}

In the analysis of the task-specific behaviour characterisations used in previous works, it can be observed that the characterisations are generally composed of relatively simple behavioural features (see Table~\ref{tab:related}). Based on these regularities, we formalise a workflow for the definition of behaviour characterisations. We show how such workflow can be used to systematically derive behaviour characterisations with minimal dependence on the experimenter's understanding of the task.

\begin{figure}
\centering
\includegraphics[width=0.9\linewidth]{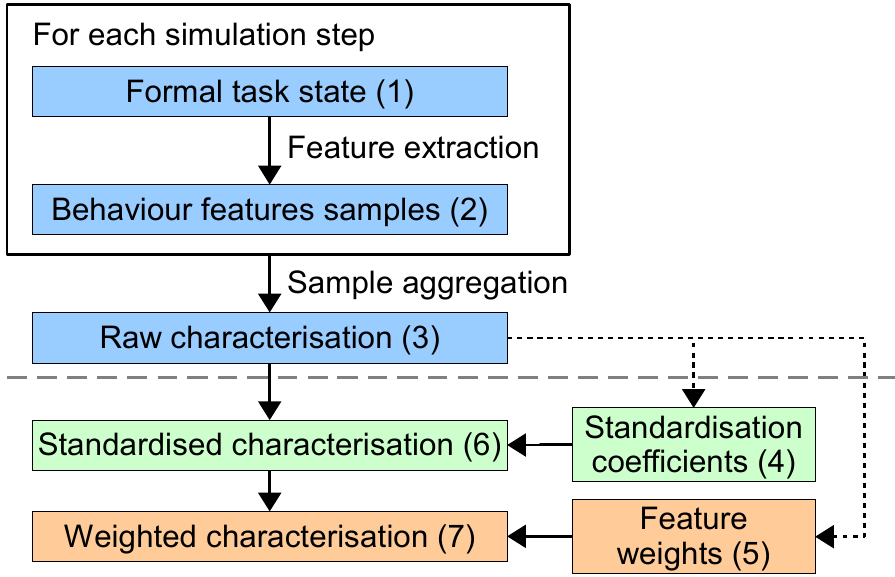}
\vspace{-5pt}
\caption{Workflow of the SDBC approach for obtaining the behaviour characterisation of each individual. The steps below the dashed line are performed after the entire population is evaluated.
See the text for details.}
\vspace{-10pt}
\label{fig:scheme}
\end{figure}

The proposed workflow is summarised in Figure~\ref{fig:scheme}. The interface between the specific task and the derivation of the behaviour characterisations is accomplished through a formal description of the current state of the task~(1) that should be specified by the experimenter. At each simulation step, a set of behaviour features is extracted from the formal task state~(2). After the simulation has ended, the behaviour feature samples are combined to assemble a fixed-length \emph{raw characterisation} vector~(3). Based on all the \emph{raw characterisation} vectors obtained in the current population, a set of standardisation coefficients is calculated~(4), and optionally, a set of feature weights~(5). The \emph{raw characterisations} are then transformed with the standardisation coefficients~(6) and feature weights~(7). The behaviour distance between individuals is then given by the Euclidean distance between the respective transformed characterisation vectors. Below, we present the task description formalism and details on each step in the workflow.


\subsection{Task state formalism}

The task state formalism separates the specific task details from the method used to devise behaviour characterisations, thus reducing the dependence on the experimenter.

The proposed formalism makes no distinction between agents and environmental features, both are treated equally as \emph{entities} of the task. Each entity $e \in E_\kappa$ is associated with a $\kappa$-tuple of state attributes ($\vartheta$), and a tuple $x$ with the constant properties of the entity. The tuple $\vartheta$ is composed of the properties of the entity that can change during task execution. To encompass multiagent tasks, we introduce the concept of \emph{entity group}. An entity group $g \in G$ can be formed by an arbitrary number of entities that share the same type and number ($\kappa$) of attributes. The cardinality ($\eta$) of a group can change during the task, and a single agent can constitute a group. The experimenter should provide the minimum ($\eta_{min}$) and maximum ($\eta_{max}$) size of each group. 


A task state is composed of (i)~a list of entity groups, and (ii)~a distance function $f_D$. The function $f_D$ should measure or estimate the physical distance between any two entities. A task state $T$ is thus defined by:
\begin{equation}
\begin{split}
T &= \left \langle g_1, \cdots, g_N, f_D \right \rangle \;,\;  g_i \in G \\
G &= \left \{ \left \langle e_1, \cdots , e_\eta \right \rangle \mid  e_i \in E_\kappa \wedge \eta \in [\eta_{min},\eta_{max}] \right \} \\
E_\kappa &= \left\{ \langle \vartheta, x \rangle \; | \; \vartheta \in \mathbb{R}^\kappa \right\} \\ 
f_D &: E_{\kappa} \times E_{\kappa'} \rightarrow \mathbb{R} \\
\end{split}
\label{eq:description}
\end{equation}


\subsection{Extraction of behaviour features}


At each time step, we automatically extract a set of behaviour features from the formal task state. These features correspond to a high-level description of the agents' behaviour and environment's state. They measure spatial relations between agents of the same entity group and between different groups, and the mean state of each group. Previous work \citep{gomes13si} has shown that averaging behaviour features over a group's agents can lead to effective and scalable characterisations for multiagent systems. We defined the following behaviour features:

\begin{itemize}[noitemsep]
\item Size of each entity group, relative to the respective limits:\\
\begin{equation}
S_g = \frac{|g| - \eta_{min_g}}{\eta_{max_g} - \eta_{min_g}} , g \in G \wedge \eta_{max_g} > \eta_{min_g}
\label{eq:feat1}
\end{equation}
\item For each entity group, mean state of the entities:\\
\begin{equation}
\vartheta_g= \left \langle \sum_{e \in g}\frac{\vartheta_{e}[1]}{|g|}, \cdots , \sum_{e \in g}\frac{\vartheta_{e}[\kappa]}{|g|} \right \rangle , g \in G
\label{eq:feat2}
\end{equation}
\item For each entity group, mean distance of each entity to the other entities of the same group (a measure of dispersion):\\
\begin{equation}
D_g = \sum_{e_i \in g}\;\sum_{e_j \in g, j \neq i}\frac{f_D(e_i,e_j)}{(|g|-1)^2} \;,\; g \in G \wedge |g| > 1
\label{eq:feat3}
\end{equation}
\item Mean pairwise distance between each two entity groups:
\begin{equation}
\begin{split}
&R = \left \langle d(g_1,g_2),\cdots,d(g_1,g_N),\cdots,d(g_{N-1},g_N) \right \rangle \\
&d(g_a,g_b) = \sum_{e_i \in g_a}\sum_{e_j \in g_b}\frac{f_D(e_i,e_j)}{|g_a||g_b|} , a \neq b 
\end{split}
\label{eq:feat4}
\end{equation}
\end{itemize}


\subsection{Aggregation of feature samples}

After a simulation has ended, the samples obtained for each behaviour feature at each time step are aggregated to assemble a fixed-length characterisation vector. Inspired by the task-specific characterisations used in previous work (see Table~\ref{tab:related}), a concatenation of the mean and final values of each behaviour feature is performed. The duration of the simulation is additionally included as a feature in the characterisation.




\subsection{Feature standardisation}

The behaviour characterisations can be composed of features that have different nature and thus have very distinct ranges. To overcome this disparity, the characterisations of the current population are standardised with the same set of coefficients at every generation. If a novelty archive exists, it also needs to be updated with those coefficients. Each behaviour feature $k$ of a vector $b$ is standardised according to:
\begin{equation}
b'_k = (b_k-\mu_k)/\sigma_k \enspace ,
\end{equation}
where $\mu_k$ and $\sigma_k$ are respectively the mean and standard deviation of feature $k$. 


\subsection{Feature weighting}

Previous work has shown that behaviour features weakly related with the task might be harmful \citep{gomes13si}. We therefore propose and evaluate a method for weighting the behaviour features according to their estimated relevance to the task. Relevance is estimated based on \emph{mutual information feature selection (MIFS)} \citep{battiti94} --- a machine learning feature selection algorithm. Mutual information is a quantitative measurement of how much one random variable tells us about another random variable. In the proposed approach, the relevance of each behaviour feature is estimated according to the mutual information between the feature values and the fitness scores of the individuals. Since the fitness score typically reflects the degree of fulfilment of a given task, this measure estimates the relevance of each behaviour feature with respect to the solution of the task.



Each behaviour feature is assigned a weight equal to the mutual information estimate plus a minimum weight~$\delta$. Therefore, the higher the mutual information, the higher the impact of that behaviour feature in the distance measure. The added minimum feature weight guarantees that none of the behaviour features is completely ignored. Each feature~$k$ of a characterisation~$b$ is weighted according to:
\begin{equation}
\label{eq:weights}
b'_k = b_k \cdot \left [ \delta + I(F;k) \right ] \enspace ,
\end{equation}
where $I(F;k)$ is an estimate of the mutual information between the fitness scores ($F$) and the behaviour feature $k$, and $\delta$ is the minimum feature weight. We set $\delta=0.25$ in all experiments, which is relatively small when compared to the typical mutual information scores of the more relevant features. Values of 0, 0.1 and 0.5 were also tested, but they yielded no improvements.

Behaviour feature weights have to be periodically updated throughout evolution. For scalability purposes, we only use the current population to calculate the weights. The computational complexity associated with weights calculation could be further reduced by performing updates less frequently, every $n$ generations, for instance.

\section{Experimental Setup}

We use three different tasks to evaluate the general applicability of our approach. The performance of SDBCs is compared with task-specific characterisations and fitness-based evolution. For the resource sharing task and predator-prey pursuit, we use task-specific characterisations that have been fine-tuned in previous works. In all tasks, the robots are homogeneous, i.e., the same neural network is copied to each robot of the group for evaluation. Each controller is evaluated in 10  simulations with randomised initial conditions.

\subsection{Gate escape task}



In this task, a group of robots must escape through a narrow gate that closes shortly after the first robot has passed. To solve the task successfully, the robots must first find the gate, and then wait for each other before starting to pass through the gate. The task is detailed in Figure~\ref{fig:setups} (top-left) and Table~\ref{tab:setup}. The fitness function $F_g$ is given by:
\begin{equation}
F_g = \frac{g + t / \tau}{1+N} \enspace ,
\end{equation}
where $g$ is the number of robots that escaped through the gate, $t$ is the simulation length, $\tau$ is the maximum simulation length, and $N$ is the total number of robots. The task-specific behaviour characterisation is a vector of length four comprising the following behavioural features (normalised to [0,1]): (i)~number of escaped robots; (ii)~time when the gate opened; (iii)~mean distance to the gate; and (iv) mean dispersion of the robots.

The task state $T_g$ is composed of three entity groups: (i)~the group of the robots that have still not passed through the gate; (ii)~the gate; and (iii)~the walls. Each robot has five attributes: its $x$ and $y$ position, turning speed, linear speed, and whether it is currently passing through the gate or not. The gate has one state variable, denoting whether it is closing or not. The surrounding walls are stateless.


\subsection{Resource sharing task}

In the resource sharing task \citep{gomes13si,gomes13gecco}, a group of robots must coordinate in order to allow each member periodical access to a single battery charging station. The energy consumption varies with the motor speed, and the charging station can only hold one robot at a time. The task is detailed in Figure~\ref{fig:setups} (top-right) and Table~\ref{tab:setup}. The fitness function $F_{s}$ is given by:
\begin{equation}
F_s = \frac{s + \overline{e}/e_{max}}{1+N} \enspace ,
\end{equation}
where $s$ is the number of surviving robots, $\overline{e}$ is the mean energy of the robots throughout the simulation, and $N$ is the number of robots. The task-specific characterisation was proposed in \citep{gomes13si}: a vector of length four, composed of: (i) the number of surviving robots; (ii) the mean energy of the robots; (iii) the mean movement speed; and (iv) the mean distance of the robots to the charging station. Each of these elements is normalised to $[0,1]$.

The task state $T_s$ is composed by two entity groups: (i)~the group of alive robots; and (ii)~the charging station. Each robot has six attributes: its $x$ and $y$ position, turning speed, linear speed, current energy level, and whether it is currently charging. The charging station has a single state variable indicating whether it is currently charging a robot or not.


\subsection{Predator-prey pursuit}

\begin{figure}[ht!]
\centering
\includegraphics[width=\linewidth]{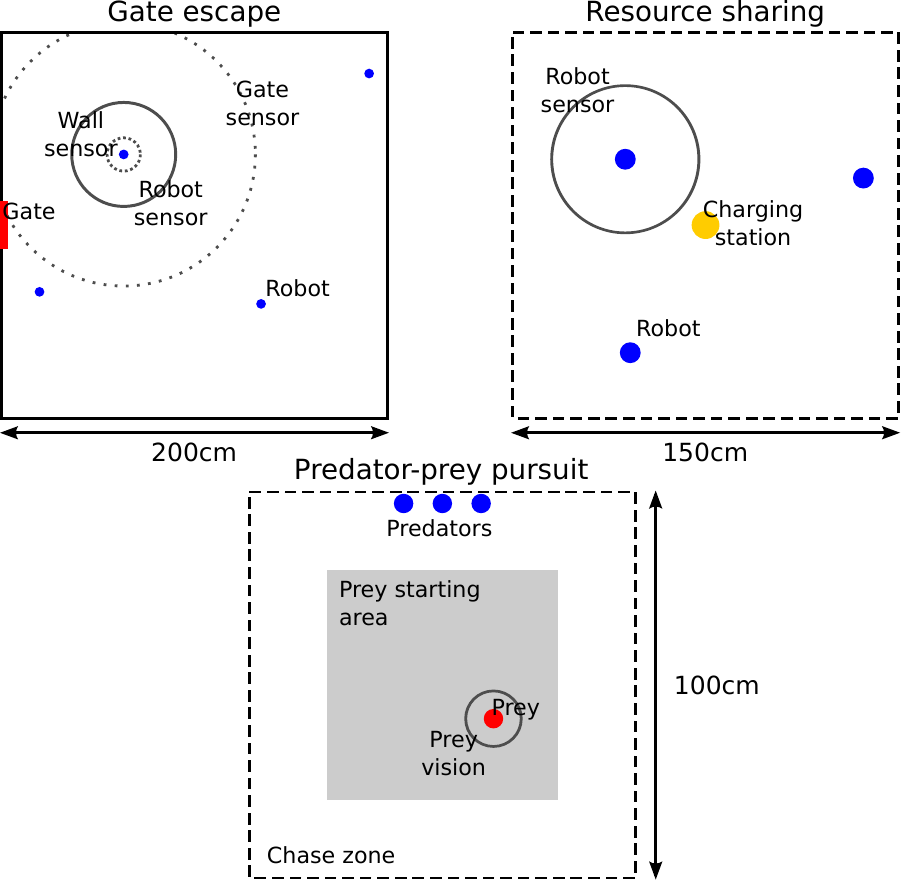}
\caption{Initial simulation conditions. In the predators-prey task, the predator's initial position  is fixed. In the other two tasks, the robots are placed randomly in the environment.}
\label{fig:setups}
\vspace{-10pt}
\end{figure}

\begin{table}[ht!]
\caption{Parameters of the tasks and the evolutionary setup.}
\begin{center}
\includegraphics[width=\linewidth]{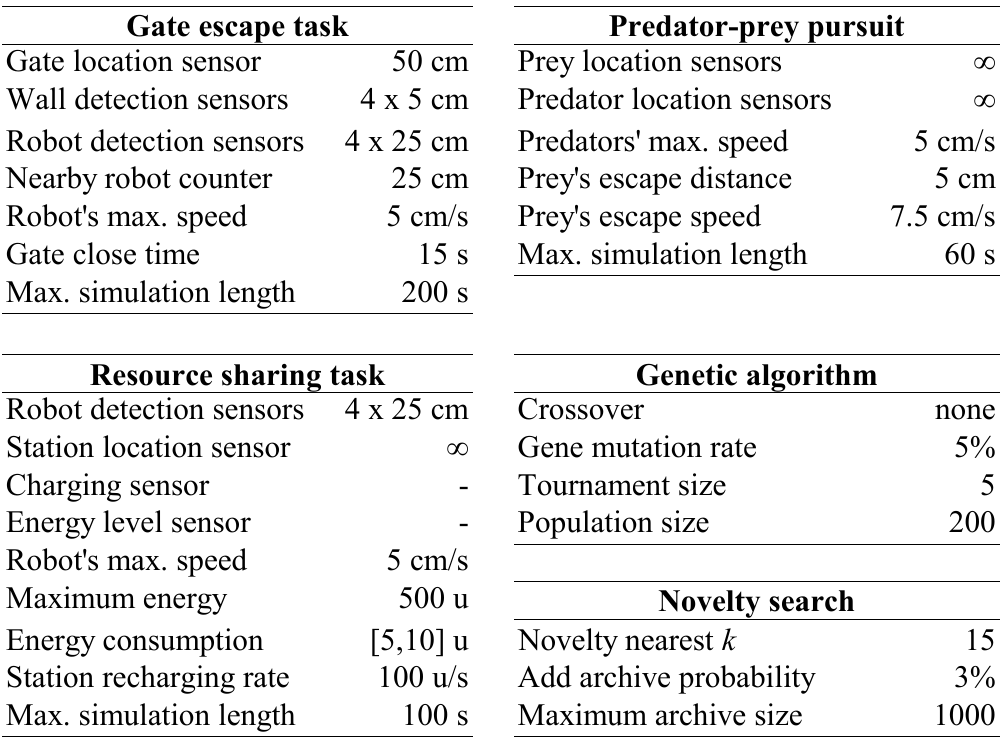}
\vspace{-20pt}
\end{center}
\label{tab:setup}

\end{table}



In the predator-prey task, three predators have the objective of capturing a single prey, i.e., one predator should physically touch the prey. The predators can sense one another. The simulation ends if the prey escapes the chase zone. Only the predators' controller is evolved, and the behaviour of the prey is preprogrammed: the prey moves away from nearby predators at full speed, and stops moving when it does not sense any predators. The task is detailed in Figure~\ref{fig:setups} (bottom) and Table~\ref{tab:setup}. The fitness function $F_{p}$ is given by:
\begin{equation}
F_{p} =
\begin{dcases*}
	2 - t/\tau & if prey captured \\
	max(d_i - d_f, 0) / size & otherwise
\end{dcases*}
\enspace ,
\end{equation}
where $t$ is the simulation length, $\tau$ is the maximum length, $d_i$ is the mean initial distance from the predators to the prey, $d_{\!f}$ is the mean final distance, and $size$ is the arena's diagonal. The task-specific characterisation \citep{gomes14} is a vector of length four, with all elements normalised to [0,1]: (i)~whether the prey was captured or not; (ii)~the simulation length; (iii)~the mean final distance of the predators to the prey; and (iv)~the mean distance of the predators to their centre of mass (dispersion) throughout the simulation.

The task state $T_p$ is composed of three entity groups: (i)~the group of the predators; (ii)~the prey; and (iii)~the chase zone boundaries. Both the predators and the prey have the same attributes: $x$ and $y$ location, turning speed, and linear speed. The environment is stateless.

\subsection{Evolutionary setup}

We used a canonical generic algorithm, where the neural network weights are directly encoded in the chromosomes. The parameters of the genetic algorithm and novelty search are listed in Table~\ref{tab:setup}. All evolutionary treatments use the same parameters. The individuals are scored with a multiobjectivisation of novelty and fitness scores: at each generation, the individuals are sorted according to their Pareto front and crowding distance \citep{deb02}, considering the fitness and novelty objectives. 



\section{Results and Analysis}

\subsection{Quality of the solutions}

\begin{figure}[tb]
\centering
\includegraphics[width=1\linewidth]{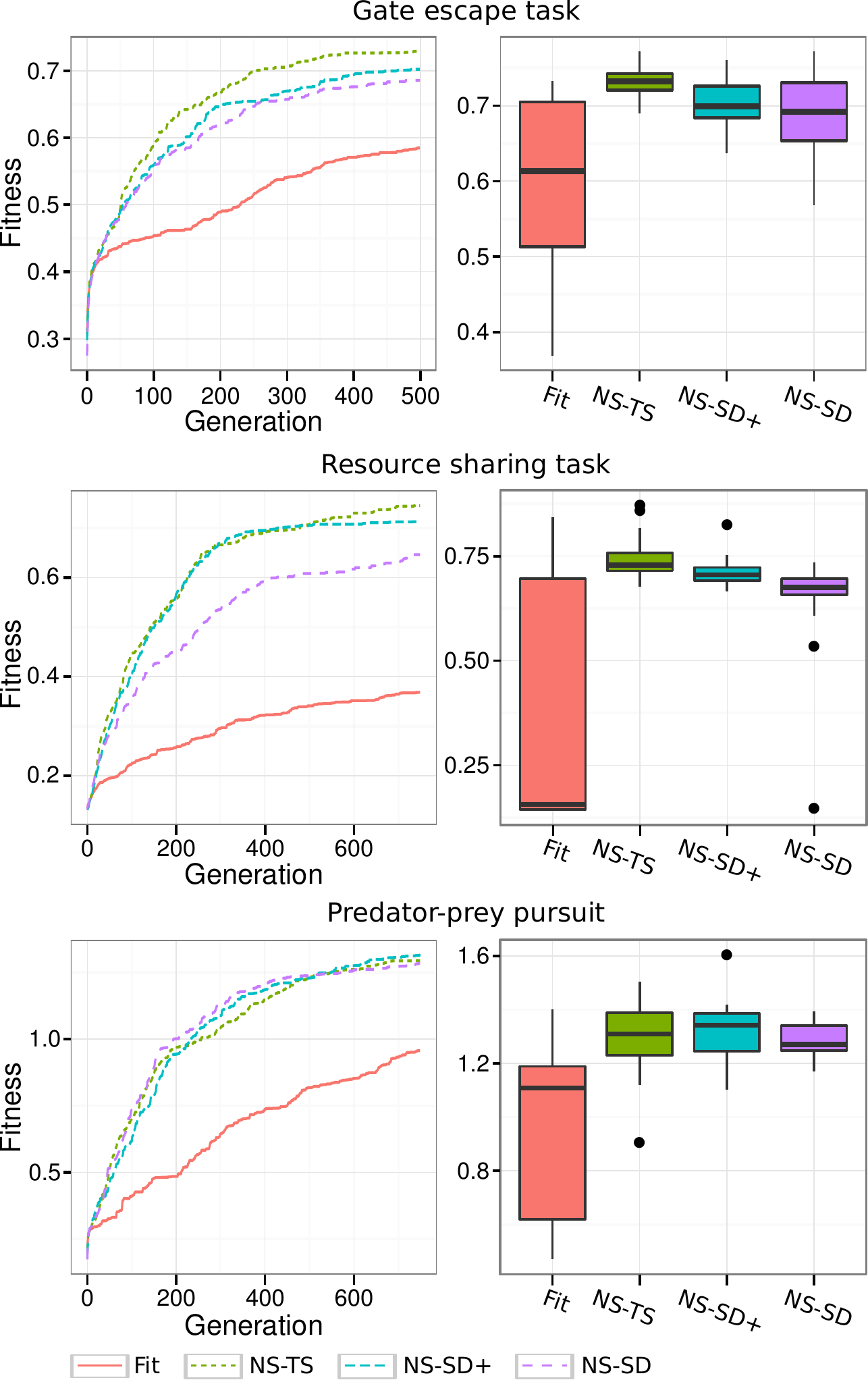}
\caption{Left: highest fitness scores found at each generation, averaged over 20 evolutionary runs for each method and task. Right: boxplots of the highest fitness scores found in each evolutionary run. The whiskers are the extreme values within 1.5\,IQR and the dots are outliers.}
\label{fig:fitness}
\vspace{-10pt}
\end{figure}

We compare fitness-driven evolution (\emph{Fit}) to novelty search with the three behaviour characterisation approaches: task-specific characterisations (\emph{NS-TS}), SDBC with feature weighting (\emph{NS-SD+}), and SDBC without weighting (\emph{NS-SD}). Figure~\ref{fig:fitness} shows the highest fitness scores achieved with each approach, in each of the three tasks.

The distribution of the highest fitness scores achieved by \emph{Fit} displays a high variance, since all tasks have some degree of deception and \emph{Fit} often converges to local optima. NS is unaffected by deception, and all variants achieve significantly higher fitness scores than \emph{Fit} (Mann-Whitney U test, $p$-value $< 0.05$). The scores achieved by \emph{NS-TS} are on average superior to \emph{NS-SD+} in the gate escape ($p$-value $=0.013$) and resource sharing tasks ($p$-value $=0.025$). The difference is, however, relatively small, and visual inspection of the solutions revealed no clear difference between the best controllers evolved by \emph{NS-TS} and \emph{NS-SD+}.



Across the three tasks, the use of feature weighting never harms the performance of SDBCs, and \emph{NS-SD+} is on average superior to \emph{NS-SD}. Significant differences between \emph{NS-SD} and \emph{NS-SD+} are, however, only present in the resource sharing task ($p$-value $< 0.01$). What makes \emph{NS-SD+} superior in this particular task is not evident, and additional experiments are needed to clarify the potential advantage provided by the weighting scheme. In the following sections, we analyse the weighting scheme and what impact it has on behaviour exploration.

\subsection{Relevance of behaviour features}

Following Equations~\ref{eq:feat1}--\ref{eq:feat4}, and the task descriptions presented in the Experimental Setup, a total of 10~behaviour features were extracted for the gate-escape task, 10~features for the resource sharing task, and 13~features for the predator-prey pursuit. After the feature sample aggregation process, this resulted in behaviour characterisations of lengths 21, 21, and 27, respectively.

Since feature extraction is systematic, it is important to determine if the behaviour features can adequately characterise the behaviour of the individuals in the context of the given task. Table~\ref{tab:weights} lists the mutual information scores of the most relevant features for each task. It can be observed that some of the behaviour features match the features defined in the task-specific characterisations (highlighted in the table with a tick). The remaining features correspond to behavioural traits that are intuitively related with the fulfilment of the task. The relevance of each feature is also consistent across the multiple evolutionary runs, as indicated by the relatively small standard deviations of the means.

\begin{table}[tb]\small
\centering
\caption{
Mutual information (MI) between behaviour features and fitness scores, averaged over each evolutionary run (\emph{Mean MI}). The standard deviation is shown in the \emph{SD MI} column. Only the features with highest MI scores are shown. \emph{(F)} marks features measured at the final state, while \emph{(M)} marks features averaged over simulation time. The right-most column indicates if a similar feature is present in the corresponding task-specific characterisation.}
\vspace{5pt}
\begin{tabular}{p{4.3cm}C{0.8cm}C{0.8cm}C{0.8cm}}
\toprule
Feature & Mean MI & SD MI & In TS char. \\ 
\cmidrule{1-4}
\multicolumn{4}{c}{\textbf{Gate escape task}}  \\ 
\midrule
Gate is closing (F) & 2.37 & 0.05 &  \\
Agent group size (F) & 1.85 & 0.03 & \checkmark \\ 
Simulation length & 1.46 & 0.04 & \checkmark \\ 
Agent group size (M) & 1.43 & 0.03 & \\ 
Gate is closing (M) & 1.39 & 0.04 & \\ 
Agent is passing gate (M) & 0.81 & 0.04 & \\ 
\cmidrule{1-4}
\multicolumn{4}{c}{\textbf{Resource sharing task}} \\ 
\midrule
Agent energy level (M) & 1.58 & 0.14 & \checkmark \\ 
Agent group size (F) & 1.56 & 0.17 & \checkmark \\ 
Agent energy level (F) & 1.12 & 0.09 & \\ 
Simulation length & 0.96 & 0.06 & \\ 
Station is occupied (M) & 0.84 & 0.08 & \\ 
Agent is charging (M) & 0.81 & 0.07 & \\ 
\cmidrule{1-4}
\multicolumn{4}{c}{\textbf{Predator-prey prey pursuit}} \\ 
\midrule
Predators-prey distance (F) & 1.47 & 0.05 & \checkmark \\
Predators-prey distance (M) & 0.98 & 0.01 \\
Predators dispersion (F) & 0.66 & 0.08 & \checkmark \\
Prey speed (F) & 0.57 & 0.03 \\
Prey-bounds distance (F) & 0.48 & 0.02 \\
\bottomrule
\end{tabular}
\label{tab:weights}
\vspace{-5pt}
\end{table}

Our results suggest that the mutual information between feature values and fitness scores is a good indicator of the relative relevance of each behaviour feature to task fulfilment. Furthermore, the analysis shows that the systematically derived characterisations include highly relevant behaviour features for the given tasks. This explains why the performance of novelty search with systematically derived characterisations is very similar to novelty search with characterisations specifically designed for each task.

\subsection{Behaviour space exploration}

\begin{figure}[tb]
\centering
\includegraphics[width=1\linewidth]{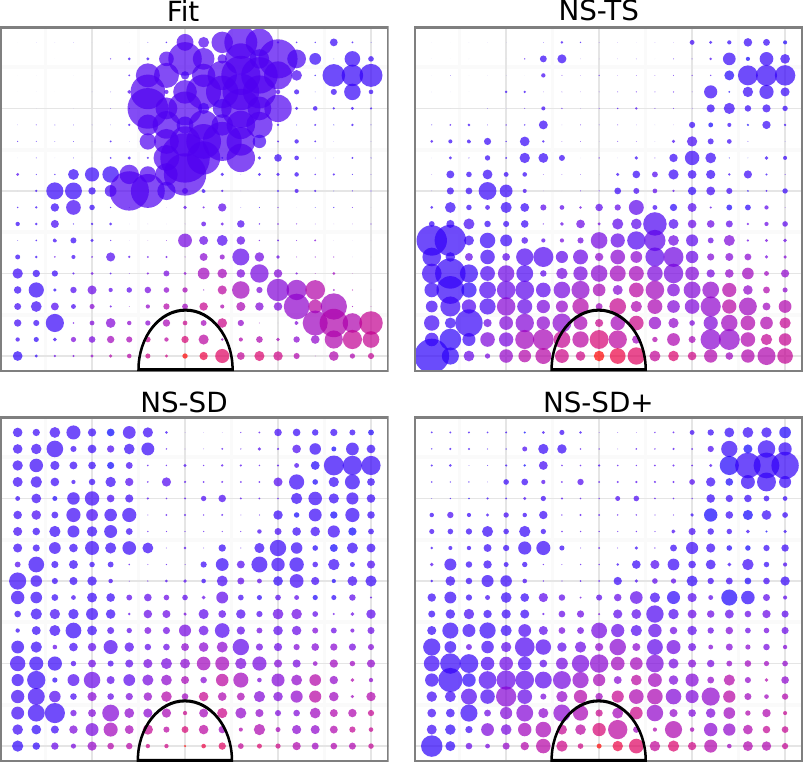}
\caption{Average behaviour space exploration in the resource sharing task. The bigger the circles, the higher the exploration in that specific behaviour region. The behaviour region associated with the highest fitness scores is highlighted with a semicircle (bottom-middle).}
\label{fig:behaviours}
\vspace{-5pt}
\end{figure}

We analysed the exploration of the behaviour space to determine the underlying differences between the novelty search variants. The behaviour space was built with the SDBC features. We used a Kohonen map to reduce the dimensionality of the space for visualisation purposes. The resulting plots for the resource sharing task can be seen in Figure~\ref{fig:behaviours}. The results for the other two tasks are similar. 

Fitness-based evolution typically explores only a narrow region of the behaviour space, which translates to premature convergence and in turn leads to an inferior performance. All novelty search variants explore the behaviour space more uniformly. The behaviour exploration in \emph{NS-SD+} follows a pattern similar to \emph{NS-TS}, but with a slightly inferior focus on the high-fitness behaviour region (highlighted in the maps). Regarding the influence of the weighting scheme, it is clear that \emph{NS-SD+} spends more effort in the high-fitness region, when compared to \emph{NS-SD}. The behaviour exploration of \emph{NS-SD} is the most uniform since there are no bias towards specific behaviour dimensions. The absence of bias causes \emph{NS-SD} to explore a behaviour region to which none of the other methods devoted significant effort (top-left corner in the maps). These results suggest that although there was no significant difference between \emph{NS-SD+} and \emph{NS-SD} in terms of solutions' quality (see Figure~\ref{fig:fitness}), the weighting scheme can actually focus behaviour exploration in the most relevant regions.


\section{Conclusion}

%

We proposed an approach for systematically deriving behaviour characterisations (SDBCs) for evolutionary robotics. The proposed approach relies on a formal description of the task state. Behaviour features are systematically extracted based on the state of the agents, their environment, and the spatial relationships between the physical entities of the task. We also proposed a feature weighting scheme that estimates the relevance of the extracted features, based on the mutual information between feature values and fitness scores. We demonstrated the proposed approaches with novelty search, using three different simulated collective robotics tasks.

Our results showed that SDBCs are on par with task-specific characterisations, with the advantage of relying less on the experimenter's task-specific knowledge. The quality of the evolved solutions is similar, and the behaviour space exploration with task-specific characterisations was similar to SDBCs with the weighting scheme. Analysing the most relevant behaviour features present in SDBCs, we could observe that they either match features of the task-specific characterisation or correspond to behavioural traits that are highly related to solving the task. While the calculated relevance scores are in accordance with our understanding of the tasks, using these scores to weight the characterisations did not translate in a significant advantage in terms of the fitness scores achieved. Our results did, however, show that feature weighting helps focus the exploration on more relevant behaviour regions and more effort is spent in regions associated with high fitness scores. 

In light of our results, we consider that the proposed systematic approach represents a promising way of defining behaviour characterisations. SDBCs reduce the dependency on the experimenter's intuition about the task, thus introducing fewer biases in behaviour exploration. Nevertheless, the systematically derived characterisations contain behaviour features that are highly related to the specific task, and can accurately capture the behaviour of the individuals. The approach is flexible and extensible, and can potentially accommodate the specific details of many tasks, possibly even outside the domain of embodied agents.

Possible extensions of the proposed approach include: (i)~extraction of additional features from the task state, e.g., by measuring inter-group relations between attributes of the same type; (ii)~improved strategies for combining the behaviour feature samples obtained during task execution, e.g., through function approximation or time discretisation; and (iii)~more elaborate weighting schemes, that could for example eliminate redundancy between behaviour features.

\subsubsection{Acknowledgements}
{\small
This research is supported by Funda\c{c}\~{a}o para a Ci\^{e}ncia e Tecnologia (FCT) grants PEst-OE/EEI/LA0008/2013, PEst-OE/EEI/UI0434/2014, SFRH/BD/89095/2012 and EXPL/EEI-AUT/0329/2013.
}



\footnotesize
\bibliographystyle{apalike}
\bibliography{alife}

\end{document}